\documentclass{llncs}
\usepackage{epsf}
\usepackage{url}
\usepackage{multirow} 

\begin{document}

\title{Relation Extraction Datasets in the Digital Humanities Domain and their Evaluation with Word Embeddings}


\author{Gerhard Wohlgenannt$^{\ast}$, Ekaterina Chernyak$^{\dagger}$, Dmitry Ilvovsky$^{\dagger}$\\Ariadna Barinova$^{\ast}$, Dmitry Mouromtsev$^{\ast}$}

\institute{$^{\ast}$International Laboratory of Information Science and Semantic Technologies\\ ITMO University, St. Petersburg, Russia \\
         $^{\dagger}$National Research University, Higher School of Economics, Moscow, Russia}

\maketitle

\begin{abstract}


In this research, we manually create high-quality datasets in the digital humanities domain
for the evaluation of language models, specifically word embedding models.
The first step comprises the creation of unigram and n-gram datasets for two fantasy novel book series for two task types each, analogy and doesn't-match.
This is followed by the training of models on the two book series with various popular word embedding model types such
as word2vec, GloVe, fastText, or LexVec.
Finally, we evaluate the suitability of word embedding models for such specific relation extraction tasks in a situation
of comparably small corpus sizes. 
In the evaluations, we also investigate and analyze particular aspects such as the impact of corpus term frequencies and task difficulty on accuracy.
The datasets, and the underlying system and word embedding models are available on github and can be easily extended with new datasets and tasks,
be used to reproduce the presented results, or be transferred to other domains.

\end{abstract}


\section{Introduction}
Language technologies are increasingly adapted in the field of digital humanities, which is
also reflected by a rising number of scientific events\footnote{For Example: \url{https://www.clarin-d.net/en/current-issues/lt4dh}}.
In this publication, we focus on the domain of literary texts, and analyze certain aspects of two well-known
fantasy novel book series, namely ``A Song of Ice and Fire'' (ASOIF) by George R.~R.~Martin, and ``Harry Potter'' (HP) by Joanne K.~Rowling.

The research question is how well current NLP methods, especially in the form of word embedding models,
perform on extracting and verifying fine-grained relations and knowledge from the comparably small book series corpora.
Firstly, we manually create high-quality datasets for the two novel book series, including various \emph{analogy} and \emph{doesn't-match} tasks
(for the precise task and dataset descriptions see Section~\ref{sec:methods}). In total the number of test units is 31362, separated into 80 task sections.
The tasks include analogies of type \emph{husband::wife}, \emph{sigil-animal::house}, \emph{geographic-entity-name::location-type}, and many others.
Secondly, we train various types of unigram and n-gram word embedding models on the book corpora, and evaluate their performance on the test datasets.
As a remark, it is not possible to apply existing generic embeddings, since the majority of entities are out of vocabulary in pre-trained models.

As most of the tasks are very hard for word embedding models trained on small corpora, we do not expect high numbers in accuracy,
but rather aim to provide easily reproducible baselines for future work on the given test datasets.
In order for other researchers to compare their methods on the test data, we make all datasets, word embeddings models and
evaluation code available online\footnote{\url{https://github.com/cicling2018-dhdata/dh-dataset}}.
By making the dataset creation and evaluation processes simple and transparent, 
we aim to provide the basis for the extension of the existing datasets, the addition of new datasets, and the enhancement of the evaluation code base.

To shed more light on the research question, in the evaluations we analyze specific aspects which influence model accuracy,
such as the impact of term frequency in the book corpus on performance, or the impact of task difficulty. 

The remainder of this paper is structured as follows: Section~\ref{sec:related} introduces related work, in Section~\ref{sec:methods} we will
describe the task types and the methods of dataset creation and dataset structure, as well as the implementation.
Section~\ref{sec:eval} presents the evaluation setup and evaluation results for the two book series, and Section~\ref{sec:concl} concludes the paper.

\section{Related Work}
\label{sec:related}

In the NLP field, word embeddings have become a very popular in recent years for language modeling and feature learning,
especially since the work of Mikolov et al.~\cite{mikolov2013w2v} on the word2vec toolkit in 2013. Word2vec includes a component to evaluate
the accuracy of its analogy feature, and an accompanying general-domain dataset. Those components inspired
parts of the datasets in the fantasy novel domain presented in this publication. Other well-known word embedding types
include GloVe~\cite{pennington2014glove}, fastText~\cite{bojanowski2016} or LexVec~\cite{SalleIV16a} --
which will be evaluated in Section~\ref{sec:eval}. In order to achieve good predictive quality, word embedding models
are usually trained on large corpora with billions of tokens. Here, we evaluate the applicability within a specialized
domain and with a small corpus.

Ghanny et al.~\cite{ghanny2016} compare different types of word embeddings, including the word2vec CBOW and skip-gram versions,
GloVe, and word2vec-f (which was trained on the output of a dependency parser).
Model performance depends on task and situation. The authors have best results with word2vec-f on some NLP tasks, 
GloVe for analogical reasoning, and word2vec for word similarity tasks. 

Wohlgenannt et al.~\cite{wohlgenannt2016lt4dh} present and evaluate methods to extract social networks using
co-occurrence statistics and word embeddings from a novel book series. 
The proposed work 
extends previous research with specific relation types like analogical reasoning and doesn't-match,
the manual creation and provision of datasets, and is applied to multiple fantasy book series.

We further relate our work to two trends in Natural Language Processing.
First of all, fantasy books recently became a popular subject of study in the Digital Humanities field
due to several factors: i) such books often have a linear timeline suitable for timeline and storyline extraction~\cite{laparra2015timelines},
b) the books feature a profound amount of direct speech for dialog~\cite{flekova2015personality} and social network analysis~\cite{bonato2016mining}.
Secondly, word embeddings and their applications are also widely studied in various fields~\cite{hellrich2016bad}, e.g.~for hyponymy detection~\cite{ustalov2017negative},
to track diachronic changes~\cite{hamilton2016diachronic} or to find corpus specific senses~\cite{kaageback2015neural}. 

\section{Methods}
\label{sec:methods}

This section includes a description of the task types of analogical reasoning and doesn't match,
gives some details on the dataset creation process, and the word embedding types used.
Finally, it provides a quick overview of the implementation and links to the online repository.

    \subsection{Task Types}
    In order to be compatible with existing tools we focus on two task types.
    Firstly, we created and evaluated a dataset for the analogical reasoning (analogy) task.
    The classical example from the original word2vec implementation is: \texttt{king} -
    \texttt{man} + \texttt{woman} = \texttt{?}, where the correct solution is \texttt{queen}.
    With embedding models this task can be simply solved with vector arithmetic. The
    word2vec toolkit contains a general domain dataset with tasks such as \\
    \emph{capital\_city$_a$, country$_a$ :: capital\_city$_b$, ?} \\
    or \emph{adjective$_a$, superlative$_a$ :: adjective$_b$, ?}. \\
    Using vector arithmetic, a candidate is selected from the whole vocabulary of the
    model, making this task quite hard due to the huge number of candidates.

    The second task type is the \emph{doesnt\_match} task according to the implementation
    within the popular Gensim library\footnote{\url{https://radimrehurek.com/gensim}}.
    Here, from an input of $n$ terms, the system has to decide which term does not belong to this list.
    This task is obviously much easier, as the system only has a small set 
    of terms (e.g.~four input terms) to choose from, and not the whole vocabulary. 


    \subsection{Dataset Creation}
    \label{meth:ds}

    In the course of this research we created two datasets each (analogical reasoning and doesn't\_match) for
    a ``A Song of Ice and Fire'' (ASOIF) and ``Harry Potter'' (HP). With the addition of n-gram datasets, the result are eight datasets.
    Inspired from the data found on the book wikis\footnote{\url{http://awoiaf.westeros.org/index.php?title=Special:Categories}}\footnote{\url{http://harrypotter.wikia.com/wiki/Main_Page}}
    we collected test data, and manually filtered, edited and extended the data to ensure high quality.
    Two domain experts each worked on the datasets for the two book series.
    Manual refinement was necessary for a number of reasons:
    i) Many of the terms in the Wiki have a very low frequency in the book text, some
    do not appear at all (but only in the extensive book appendices of ASOIF or related books by the same author).
    ii) We removed ambiguous terms from the datasets, as currently the focus of the datasets is not word sense disambiguation, but relation modeling.
    As an example, in the ASOIF world, ``Nymeria'' is the name of Arya's direwolf, but also the first name of Nymeria Sand (a character).
    iii) Additionally to our original unigram models and datasets, to properly capture some of the entities, it was necessary to 
    also provide n-gram models and datasets. The selection of proper surface forms for n-grams, 
    and the assignment to the unigram or n-gram dataset also required some manual intervention.

    \subsection{Word Embedding Models}

    To address the tasks defined in the dataset, obviously, many techniques and combinations of methods
    are feasible. At the current stage, we focus on the application of word embedding models.
    Word embedding models have been shown to be very successful on many NLP tasks~\cite{ghanny2016}, and furthermore they
    are easy to train, apply and compare.

    In the next section we evaluate various word embedding models based on these four well-known model types:
    \paragraph{word2vec:} Word2vec~\cite{mikolov2013w2v} was developed by a team of researchers at Google. It applies two-layer neural networks which are trained to reconstruct the linguistic context of words.
        In training, the input is a (typically large) corpus, the results are the word embeddings. Every word (above the threshold frequency) is assigned a dense vector of floating point values,
        usually in the range of 50--300 dimensions. Proximity in vector space reflects similar contexts in which words appear. word2vec includes two model architectures: CBOW and skip-gram.
        With CBOW, the model predicts the current word by using a window of surrounding words. Using skip-gram, the model predicts the surrounding context of the current word.

    \paragraph{GloVe:} GloVe~\cite{pennington2014glove} takes a different route to learning continuous vector representations of words. GloVe applies dimension-reduction on a word-word co-occurrence matrix.
        The model is trained to learn word vectors such that their dot product equals the logarithm of the words' probability of co-occurrence.

    \paragraph{fastText:} FastText~\cite{bojanowski2016} is based on the skip-gram model, but also makes use of word morphology information in the training process.
        By using sub-word information, fastText can also supply vectors for out-of-vocabulary words.

    \paragraph{LexVec:} Finally, LexVec~\cite{SalleIV16a} uses a weighted factorization of the Positive Point-wise Mutual Information (PPMI) matrix via stochastic gradient descent.
        It employs high penalties for errors on frequent co-occurrences, and performs very well on word similarity and analogy in the experiments by the authors. 


    \subsection{Implementation}

    The implementation has two main components, the creation of datasets and
    the use of those datasets to evaluate the models. As stated, all code, the word embedding
    models, and the datasets are available online\footnote{\url{https://github.com/cicling2018-dhdata/dh-dataset}}.
    The results presented here can be reproduced by cloning the repository and running the evaluation
    scripts, and all parts (datasets, models, etc.) can be extended easily.

    Regarding the creation of datasets, the system uses a simple format to define the
    task units for any section in the dataset, e.g.~all the \emph{child::father} relations.
    The data format is documented in the github repository. From the definitions,
    the \texttt{create\_questions.py} script creates the evaluation dataset as all permutations of the input definitions.

    For the evaluation of the system, there are two main scripts for the \emph{analogies} and \emph{doesn't-match}
    tasks. Both first iterate over the word embedding models defined in the configuration file,
    run the task units from the datasets, and obviously collect and aggregate all the evaluation results.
    The implementation makes use of the Gensim library~\cite{gensim} for loading the models, and performing
    the two basic task types.
    For details on script usage and dataset extension see the system documentation online on github.

\section{Evaluation}
\label{sec:eval}
The evaluation section first describes specifics of the text corpora, the model settings and the datasets,
and then provides the evaluation results for analogical reasoning and the doesn't-match tasks.

\subsection{Evaluation Setup}

    \subsubsection{Text Corpora}
    As already mentioned, the experiments were run on the plain text corpora of ``A Song of Ice and Fire'' (ASOIF) by George R.~R.~Martin (books 1-4),
    and ``Harry Potter'' (HP) by Joanne K.~Rowling (all books).
    The ASOIF corpus has a size of 6.9M and contains about 1.3M word tokens. The HP series is of similar size, with 6.5M file size and 1.1M tokens.
    Corpus preprocessing only consisted of the removal of punctuation, quotation marks and newline characters. In order to support also 
    datasets for n-grams, we applied the word2phrase tool from word2vec to create the n-gram corpus versions.

    \subsubsection{Models Trained}

        To compare and evaluate the performance of various word embedding types, we trained the following
        models on the two corpora (all available on github): 

        \begin{description}
            \item{\emph{w2v-default:}} This is a word2vec model trained with the default settings: 200 vector dimensions,
            skip-gram, word window size:5, etc\footnote{-cbow 0 -size 200 -window 5 -negative 0 -hs 1 -sample 1e-3 -threads 12}.
            \item{\emph{w2v-ww12-300-ns:}} window size of 12, 300-dim., negative sampling.
            \item{\emph{w2v-CBOW:}} same settings as \emph{w2v-ww12-300}, but CBOW instead skip-gram.
            \item{\emph{GloVe:}} defaults (window size 15). Except: 200-dim vectors instead 50-dim.
            \item{\emph{fastText:}} We used the default settings, except: 25 epochs, window size of 12
            \item{\emph{LexVec:}} We used the default settings, except: 25 epochs, window size of 12 
        \end{description}

        The models and test dataset are available in versions for unigrams and n-grams.
        The n-gram data can be easily recognized by the suffix ``\texttt{\_ngram}'' as part of the respective filenames.
        All models are available online in the github repository.

    \subsection{Dataset Description}
        The dataset creation process is described in Section~\ref{meth:ds}. The resulting datasets are:

        \begin{description}
            \item{\emph{questions\_soiaf\_analogies.txt:}} Analogical reasoning tasks for the ASOIF series. Contains 8 different task types with a total of 2848 tasks.
            \item{\emph{hp\_soiaf\_analogies.txt:}} Analogical reasoning tasks for the Harry Potter series. 17 different task types and 4790 tasks in total.
            \item{\emph{questions\_soiaf\_doesnt\_match.txt:}} Doesn't-match tasks for ASOIF, with 13 sections, and 11180 total tasks.
            \item{\emph{questions\_hp\_doesnt\_match.txt:}} Doesn't-match tasks for the HP series, with 19 sections, and 8340 total tasks.\\
            Additionally to these four datasets, there exist four more datasets for n-grams, with similar filenames, but including
            the marker ``\texttt{\_ngram}''. The n-gram datasets include in total 4204 task units in 23 sections. 
        \end{description}


    \begin{table}[t]
    \caption{ASOIF analogy dataset: Accuracy of various word embedding models on various selected analogy task types, and total accuracy.}
    \begin{center}
    \begin{tabular}{lllllll}
    \hline

      \multirow{2}{*}{Task Type} & first-name-~~   & child-~~ & husband-~~& geo-name-~~   & houses-~~ & \multirow{2}{*}{Total}~~ \\ 
                                 & last-name~      & father~~ & wife      & location~     & seats~    &                          \\ \hline
 
        Number of tasks:         & 2368   & 180  & 30   & 168   & 30    & 2848  \\ \hline

        w2v-default              &  20.73 & 3.33 & 6.67 & 1.19  & 43.33 & 18.43  \\ 
        w2v-ww12-300-ns          &  39.53 & 1.67 & 0.0  & 10.71 & 26.67 & 34.38  \\ 
        w2v-CBOW                 &   0.46 & 6.11 & 10.0 & 7.14  & 6.67  & 1.37  \\ 
        GloVe                    &  40.58 & 6.11 & 3.33 & 1.19  & 26.67 & \textbf{34.8}  \\ 
        fastText                 &  37.67 & 4.44 & 0.0  & 13.1  & 26.67 & 32.94  \\ 
        LexVec                   &  40.08 & 3.89 & 6.67 & 0.0   & 23.33 & 34.38  \\

    \end{tabular}

    \label{tab1a}
    \end{center}
    \end{table}

    \begin{table}[h]
    \caption{Harry Potter analogy dataset: Accuracy of various word embedding models on various selected analogy task types, and total accuracy.}
    \begin{center}
    \begin{tabular}{llllll}
          \hline

      \multirow{2}{*}{Task Type} & first-name-~~   & child-~~ & husband-~~&    name-~~    & \multirow{2}{*}{Total}~~ \\ 
                                 & last-name~      & father~~ & wife      &    species~~  &                          \\ \hline

        Number of tasks:       & 2390               &     224      &       72     &      566       & 4790  \\ \hline

        w2v-default            &  21.51 & 11.61 & 11.11 & 21.02 & 21.82  \\ 
        w2v-ww12-300-ns        &  50.46 & 11.16 & 43.06 & 32.69 & 34.11  \\ 
        w2v-CBOW               &  2.13  & 6.7  & 0.0    & 9.36  & 4.59  \\ 
        GloVe                  &  43.68 & 6.7  & 22.22  & 25.27 & 30.96  \\ 
        LexVec                 &  43.97 & 8.04 & 37.5   & 36.93 & \textbf{34.72}  \\ 
        fastText               &  38.33 & 3.57 & 33.33  & 13.25 & 23.88  \\ 

    \end{tabular}
    \label{tab1b}
    \end{center}
    \end{table}

\subsection{Analogy Task Results}

    As mentioned, in the analogy task, the input is a triple of terms, which can be read as $x_1$ is to $x_2$, what $y_1$ is to $y_?$.
    For example, \emph{man} is to \emph{king} what \emph{woman} is to $X$. Or, in ASOIF, \emph{kraken} is to \emph{Greyjoy} what \emph{lion} is to $X'$ (correct: \emph{Lannister}).
    The evaluated system has to guess the correct answer from the whole vocabulary. The vocabulary size in the models used here is between $11K$ to $60K$ terms.
    Given the comparably small corpus size, and the ambiguity of relations between terms, this task is very hard.

    Every task is defined by four terms, three input terms, and the correct answer.
    The tasks are split into various sections, for example predicting \emph{child-to-father} relations, \emph{houses-to-their-seats}, etc.
    The ASOIF and HP datasets contain in total 7638 task units. For every unit, Gensim uses vector arithmetic to
    calculate the candidate term, and compares it to the correct solution given in the dataset.

    Table~\ref{tab1a} presents the evaluation results for the ASOIF analogy dataset.
    Due to space limitations, we selected the results for five dataset sections, and
    the aggregated results. Best results were provided by \emph{GloVe}, but there is no
    considerable difference to word2vec variants like \emph{w2v-ww12-300-ns} and to \emph{LexVec}.
    Both the \emph{w2v-default} setting with its small word window of 5, and especially the word2vec \emph{CBOW}
    method are unsuited for the task. Generally the accuracy is low with ca.~34\%,
    reasons for the difficulty of the task setting are given above.

    The evaluation of analogical reasoning with the dataset for the HP books confirms the observations made on the ASOIF dataset.
    Here, the setting \emph{LexVec} performs best, followed by \emph{w2v-ww12-300-ns} and \emph{GloVe}.  Again, the score of \emph{w2v-CBOW} is extremely low. 
    The data suggests that for the analogy task with such a small corpus a large word window is helpful to counter data sparsity.

    As a general remark, it was challenging to create a large number of high-quality relations for analogical reasoning.
    Partly because of ambiguity problems, as there are many re-occurring names and nicknames of entities in the large ASOIF universe.
    Furthermore, relations over change over time, for example in the ASOIF series, the character \emph{Jon Snow} is first wrongly thought to be a bastard son of \emph{Ned Stark}, 
    which is later revealed as not true.

\subsection{Doesn't\_match  Task Results}


    In the doesn't-match task, the input dataset contains four terms, where three are semantically connected, and one intruder is mixed in.
    To distinguish the correct answer, it is explicitly provided in the dataset after a term separation symbol.
    The evaluation script calls Gensim to provide the intruder candidate,
    which is then compared to the correct answer.  Internally, Gensim computes the mean vector from all input vectors, and
    then calculates the cosine distances. The vector with the largest distance is selected as not matching the rest.
    The random baseline for this task is $\frac{1}{4}$ (0.25).

    \begin{table}[ht]
    \caption{ASOIF doesnt\_match dataset: Accuracy of various word embedding models on selected doesn't\_match task types, and total accuracy.}
    \begin{center}
    \begin{tabular}{llllll}
          \hline
      \multirow{2}{*}{Task Type} & family-     & names-of-~~ & Stark~~  & free     & \multirow{2}{*}{Total}~~ \\ 
                                 & siblings~~  & houses      & clan~~   & cities~~ &                          \\ \hline

            Number of tasks:       & 160    & 7280  & 1120  & 700   & 11180  \\ \hline

            w2v-default            &  85.63 & 65.32 & 94.29 & 90.43 & \textbf{74.54} \\ 
            w2v-ww12-300-ns        &  85.63 & 53.7  & 88.39 & 91.86 & 66.85  \\ 
            w2v-CBOW               &  78.13 & 50.25 & 84.29 & 85.14 & 62.21  \\ 
            GloVe                  &  80.63 & 67.69 & 89.64 & 88.71 & 73.28  \\ 
            fastText               &  84.38 & 54.3  & 84.64 & 91.14 & 66.86  \\ 
            LexVec                 &  80.0  & 56.92 & 81.96 & 90.43 & 67.51  \\ 

    \end{tabular}
    \label{tab2a}
    \end{center}
    \end{table}

    Table~\ref{tab2a} gives an overview of the results on the $11180$ task units defined for ASOIF.
    Due to the lower task complexity accuracy numbers are around 75\%, with best results for
    \emph{w2v-default} and \emph{GloVe}. 
    Both for ASOIF and for Harry Potter (Table~\ref{tab2b}) \emph{w2v-default} performs surprisingly well on the doesn't-match tasks,
    which suggests that for this task type (semantic word similarity) a smaller and more focused word context has benefits.

    \begin{table}[ht]
    \caption{HP doesnt\_match dataset: Accuracy of various word embedding models on selected doesn't\_match task types, and total accuracy.}
    \begin{center}
    \begin{tabular}{llllll}
          \hline
      \multirow{2}{*}{Task Type} & family-     & Gryffindor-~~ & magic-~~    & wizards-  & \multirow{2}{*}{Total}~~ \\ 
                                 & members~~   & members       & creatures~~ & animagi~~ &                          \\ \hline

         Number of tasks:       & 440 & 2800 & 700 & 200 & 8340  \\ \hline

         w2v-default            &  85.23 & 82.07 & 59.86 & 72.0 & \textbf{71.16}  \\ 
         w2v-ww12-300-ns        &  88.18 & 73.18 & 33.57 & 56.0 & 66.38  \\ 
         w2v-CBOW               &  56.59 & 53.36 & 40.71 & 52.0 & 55.78  \\ 
         GloVe                  &  84.09 & 69.46 & 35.57 & 53.0 & 65.49  \\ 
         fastText               &  85.68 & 76.71 & 35.14 & 34.5 & 64.84  \\ 
         LexVec                 &  83.86 & 69.5  & 39.71 & 60.0 & 61.92  \\ 

    \end{tabular}
    \label{tab2b}
    \end{center}
    \end{table}

    The Harry Potter dataset defines a total of 8340 task units in 19 sections.
    The evaluation results are in line with the ASOIF evaluations, with the best results for \emph{w2v-default} with an accuracy
    of $71.16\%$, followed by other word2vec variants, GloVe, and fastText.
    Again, results per task section vary considerably. Section~\ref{sec:eval:freq} analyzes the influence of term frequencies
    in the corpus on accuracy, and section~\ref{sec:eval:diff} discusses the impact of task difficulty.
    Our results also confirm Ghanny et al.~\cite{ghanny2016}, with good results for GloVe on analogical reasoning, and 
    word2vec for word similarity.

\subsection{Results for N-Gram Datasets and Models}
\label{sec:ngrams}

    Additionally to the unigram data, we also trained n-gram models and created n-gram datasets. 
    For the ASOIF book series, we define 192 task units for analogical reasoning and 2000 units for the doesn't-match task,
    the respective numbers for HP are 92 and 1920.
    As the results of the n-gram evaluations are in many aspects in line with unigram data, we will keep the analysis short and focus on differences to the unigram results.
    Performance on analogical reasoning is generally lower, with only about 10-15\% accuracy in the n-gram setting. 
    This might (in part) be caused by lower term frequencies associated with n-grams. 
    For the doesn't-match task, accuracy is similar to the unigram evaluation, with the exception of fastText embeddings. 
    As fastText uses subword information, it performs very well on n-grams which share lexical elements. 
    For the ASOIF dataset, fastText provides $92.1$ accuracy for the doesn't-match task.

\subsection{Impact of Term Frequencies}
\label{sec:eval:freq}

We investigate the correlation between the frequency of the terms in the respective book corpus, and the correct guesses 
in the doesnt-match task. 
In general, the expectation was that a higher frequency of terms in the book leads to a ``better'' word embeddings vector
for the term, and therefore to higher results. Our experiments confirm this expectation only in part.
As every task unit consists of four terms (the three that are related, and the intruder to be found),
the question is which of the term frequencies to use. Table~\ref{corr1} shows the \emph{correlation scores} 
for different aspects of term frequency: (i) the frequency of the real \emph{intruder} which had to be found,
(ii) the frequency of the term chosen as intruder using the model, (iii) the frequency bin of the chosen intruder, 
and (iv) the average frequency of all four terms in the task unit.

    \begin{table}[ht]
    \caption{Correlations of term frequencies with accuracy in tasks for unigram and n-gram models trained on \emph{Harry Potter} and \emph{A Song of Ice and Fire}}
    \begin{center}
    \begin{tabular}{lllllll}
          \hline

    \multicolumn{2}{c}{\textbf{Correlation of}} &  ~Freq. of       &  ~Freq. of chosen &  ~Freq. bin of     &  ~Avg. Freq.     & ~\multirow{2}{*}{Difficulty} \\
    \multicolumn{2}{c}{\textbf{accuracy to}}    &  ~real intruder  &  ~intruder        &  ~chosen intr.     &  ~of task terms  &                             \\ \hline

        \multirow{2}{*}{ASOIF}        & unigram & ~-0.10           & ~0.31         &  ~0.29             & ~0.05            & ~0.30   \\ 
                                      & n-gram  & ~0.15            & ~0.17         &  ~0.18             & ~0.12            & ~0.10   \\ \hline 
                  \multirow{2}{*}{HP} & unigram & ~-0.03           & ~0.05         &  ~0.20             & ~0.02            & ~0.18   \\ 
                                      & n-gram  & ~-0.25           & ~0.31         &  ~0.58             & ~-0.19           & ~0.25   \\ \hline
   \multicolumn{2}{c}{\textbf{Average results}} & ~-0.06           & ~0.21         &  ~0.31             & ~0.00            & ~0.21   \\ \hline

    \end{tabular}
    \label{corr1}
    \end{center}
    \end{table}

The data suggests that the intruder is not easier to find if it has a higher frequency in the books
(column ``frequency of real intruder'' in Table~\ref{corr1}). 
One reason might be that very frequent entities often change their context as the story progresses. E.g.,
in ASOIF the \emph{Arya} character first lives as child with her family, then moves into the context of the capital city,
later travels the country with changing companions, and finally ends up to be trained as assassin on a different continent, 
so the context changes are drastic. 
More research is needed to investigate the phenomenon of little influence of the frequency of the intruder.

On the other hand, the frequency of the intruder guessed using the model is correlated with accuracy,
the level of correlation is similar to the correlation between task difficulty and accuracy. To further analyze
this finding, we created frequency bins and measured the accuracy per bin, see Table~\ref{bin1} for results for the LexVec models.

    \begin{table}[ht]
    \caption{Accuracy of \emph{doesn't match} tasks depending on the frequency of the term suggested by the word embedding model. Evaluated using LexVec embeddings.}

    \begin{center}
    \begin{tabular}{lllllll}
          \hline
        \multicolumn{2}{c}{\textbf{Correct Results}} &  Bin 1       &  Bin 2       &  Bin 3        &  Bin4        & Bin 5         \\

         \multirow{2}{*}{ASOIF}        & unigram     & ~0.29 (140)  & ~0.82 (694)  &  ~0.40 (2031) & ~0.63 (5262) & ~0.92 (3053)  \\ 
                                       & n-gram      & ~0.61 (431)  & ~0.57 (349)  &  ~0.84 (268)  & ~0.85 (420)  & ~1.00 (532)   \\ \hline 
         \multirow{2}{*}{HP}           & unigram     & ~0.32 (800)  & ~0.64 (1033) &  ~0.47 (1776) & ~0.71 (3293) & ~0.73 (1438)  \\ 
                                       & n-gram      & ~0.13 (676)  & ~0.19 (321)  &  ~0.28 (479)  & ~0.89 (131)  & ~0.97 (313)   \\ \hline
        \multicolumn{2}{c}{\textbf{Average results}} & ~0.337       & ~0.555       &  ~0.525       & ~0.77        & ~0.905        \\ \hline

    \end{tabular}
    \label{bin1}
    \end{center}
    \end{table}

Table~\ref{bin1} presents the evaluation results depending on the frequency bin of the intruder suggested by the model.
Terms are categorized as follows: Bin 1 contains the terms occurring in the books less than 20 times, bin 2 is for terms
with 20--50 occurrences, bin 3 for frequency 50--125, bin 4 for frequency 125--500, and bin 5 for term frequencies above 500.
The table shows the ratio of correct guesses in the \emph{doesn't match task} for the HP and ASOIF book series, with the 
number of tasks units per bin given in parenthesis.
The general tendency is that term frequency of the suggested intruder impacts accuracy, but the evaluations show high variability.

\subsection{Impact of Task Difficulty}
\label{sec:eval:diff}

Finally, we analyze the impact of task difficulty on accuracy.
In the \emph{doesn't match}-task, task difficulty, can be controlled via the level of similarity of the mixed-in term to the other three terms.
To this end, in the datasets we define 20 intruding terms per doesn't match triple, which are grouped into four 
difficulty categories (1--4), with five terms per category. In the hardest category mixed-in terms are semantically very close and of same syntactic type or named entity class, 
whereas in the easiest category, there is no semantic relation of the intruder to the doesn't match triple.
In the ASOIF dataset for example, in section ``family-siblings'' for the triple of siblings \emph{Jaime Tyrion Cersei}, in the hardest category,
intruding terms are e.g. \emph{Tywin} (father) or \emph{Joffrey} (son of the siblings). And in the easiest category, an example of an intruding term is \emph{dragon} (not related).

    \begin{table}[ht]
    \caption{Accuracy of ASOIF and HP datasets in \emph{doesn't-match} tasks depending on task difficulty.}
    \begin{center}
    \begin{tabular}{lcccc|c}
          \hline
      & \multicolumn{5}{c}{Task Difficulty} \\
      Datasets / Models & ~1 (hard) & ~2 (rather hard) & ~3 (rather easy)  & ~4 (easy)~ & ~Total~ \\  \hline
             ASOIF GloVe            & 54.28 & 63.32 & 86.26 & \textbf{89.27} & 74.54 \\ 
             HP LexVec              & 47.62 & 68.29 & 81.91 & \textbf{86.81} & 71.16 \\ 
             ASOIF Ngram fastText   & 67.00 & 74.60 & 85.00 & \textbf{93.60} & 80.05 \\ 
             HP Ngram w2v-default   & 48.54 & 72.01 & 87.08 & \textbf{95.83} & 75.89 \\ 
    \end{tabular}
    \label{tab:diff}
    \end{center}
    \end{table}

Table~\ref{tab:diff} presents the results of the doesn't-match tasks based on task difficulty levels, showing the results for unigram and n-gram datasets of selected embedding models.
The impact of task difficulty on accuracy is clearly evident from the data, for the hardest task category accuracy is around 50\%, which leaves a lot of room for future work.

\section{Conclusion}
\label{sec:concl}

In this publication, we introduce datasets in the digital humanities domain for evaluating word embedding and relation extraction systems,
and apply various embedding methods to provide benchmark evaluations. The datasets include
31362 analogical reasoning and doesn't-match relations for the ``A Song of Ice and Fire'' (by G.R.R. Martin)
and ``Harry Potter'' (by J.K. Rowling) book series.

The contributions of this work are as follows: (i) the manual creation of high-quality datasets
in the digital humanities domain, (ii) providing models for the two book series trained with various
word embedding techniques, (iii) evaluating the suitability of word embeddings trained on
small corpora for the tasks at hand, including the evaluation of unigram and n-gram tasks,
and analyzing the impact of task difficulty and corpus term frequencies on accuracy, 
and finally (iv) the provision of the code-base and related resources, 
which are easy to extend with new task types, test data, and models.

A very interesting direction of future research is leveraging multimodal data (images, videos, etc.) available in abundance in the given domain
for the creation of entity representations. Thoma et al.~\cite{thoma2017} propose multimodal embeddings which combine embeddings
learned from text, images, and knowledge graphs.
Furthermore, we plan to apply crowdsourcing on a subset of task units to get an estimate of human level accuracy on the various task types and difficulty levels.

\section*{Acknowledgements}
This work was supported by the Government of the Russian Federation (Grant 074-U01) through the ITMO Fellowship and Professorship Program.
Furthermore, the article was prepared within the framework of the Basic Research Program at the National Research University Higher School of Economics (HSE) 
and supported within the framework of a subsidy by the Russian Academic Excellence Project '5-100'. It was supported by the RFBR grants 16-29-12982, 16-01-00583.

%


\bibliographystyle{splncs}
\bibliography{cicling}
\end{document}